\documentclass{article}

\usepackage{PRIMEarxiv}

\usepackage{multirow}
\usepackage{caption}
\usepackage{subcaption}
\usepackage{amsmath} 
\usepackage[utf8]{inputenc} 
\usepackage[T1]{fontenc}    
\usepackage{url}            
\usepackage{booktabs}       
\usepackage{amsfonts}       
\usepackage{nicefrac}       
\usepackage{microtype}      
\usepackage{lipsum}
\usepackage{fancyhdr}       
\usepackage{graphicx}       

\usepackage[pagebackref,breaklinks,colorlinks]{hyperref}

\graphicspath{{media/}}     

\title{V-MAD: Video-based Morphing Attack Detection \\ in Operational Scenarios
}

\author{
  Guido Borghi, Annalisa Franco, Nicolò Di Domenico, Matteo Ferrara, Davide Maltoni \\
  Department of Computer Science and Engineering \\
  University of Bologna \\
  \texttt{\{name.surname\}@unibo.it} \\
}

\begin{document}
\maketitle

\begin{abstract}
In response to the rising threat of the face morphing attack, this paper introduces and explores the potential of Video-based Morphing Attack Detection (V-MAD) systems in real-world operational scenarios. 
While current morphing attack detection methods primarily focus on a single or a pair of images, 
V-MAD is based on video sequences, exploiting the video streams often acquired by face verification tools available, for instance, at airport gates. 
Through this study, we show for the first time the advantages that the availability of multiple probe frames can bring to the morphing attack detection task, especially in scenarios where the quality of probe images is varied and might be affected, for instance, by pose or illumination variations.
Experimental results on a real operational database demonstrate that video sequences represent valuable information for increasing the robustness and performance of morphing attack detection systems. 
\end{abstract}

\section{Introduction}

In the last decades, the wide diffusion of Facial Recognition Systems (FRSs)~\cite{oloyede2020review} has significantly increased the demand for robust security measures to counter emerging threats, including those associated with the face morphing attack~\cite{FerraraFM14, Ferrara2022} through which it is possible to create a sort of hybrid face with a double identity.

Current methods to counter this kind of attack are referred to as Morphing Attack Detection (MAD) systems~\cite{scherhag2019face,scherhag2017biometric} and predominantly are focused on the analysis of the single document image, \textit{i.e.,} Single-image Morphing Attack Detection (S-MAD)~\cite{hamza2022generation,borghi2023revelio} or pairs of images (the document and the live acquired ones), \textit{i.e.,} Differential Morphing Attack Detection (D-MAD)~\cite{scherhag2020,borghi2021double}. 
However, in real-world operational scenarios such as Automated Border Control (ABC) gates in international airports \cite{del2015face}, many commercial FRS technologies often acquire video streams, providing a continuous sequence of frames~\cite{del2016automated}. 
This operational mode is indeed considered in the evaluation of FRS vulnerability to morphing attacks: the Morphing Attack Potential~\cite{MAP} metric is defined considering multiple verification attempts.
On one hand, the presence of multiple frames could strengthen the morphing attack, increasing the probability of success, since a single match for a frame of the sequence might be sufficient to pass the verification check~\cite{frontex2010automated}. On the other hand, we believe that the use of multiple frames could be advantageous from the MAD task perspective and must be considered.

\begin{figure}
    \centering
    \includegraphics[width=0.6\linewidth]{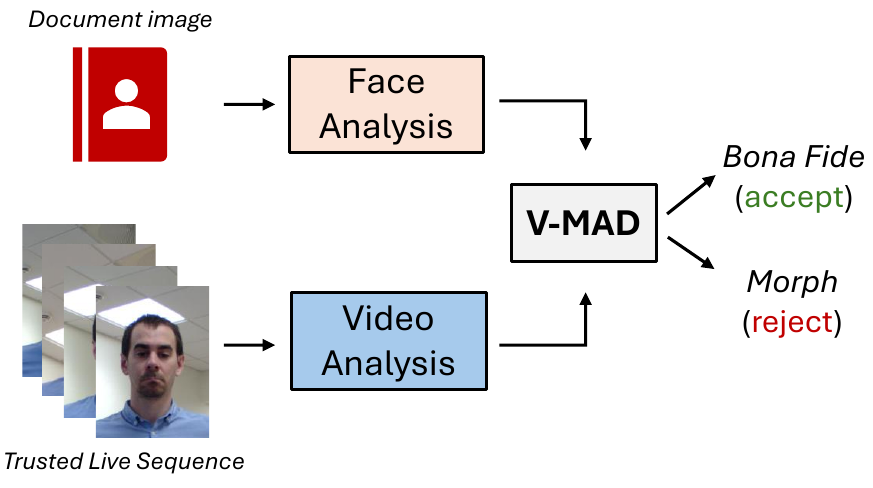}
    \caption{In operational scenarios, V-MAD represents a viable paradigm when a single probe document image is compared with an input sequence to detect whether the image is morphed or not. V-MAD differs from currently available literature solutions, based only on a single (S-MAD) or a pair (D-MAD) of images. }
    \label{fig:eye}
\end{figure}

Therefore, in this paper, we introduce the \textbf{Video-based Morphing Attack Detection} (V-MAD) task, as an effective solution for adapting MAD algorithms to real-world operational scenarios, such as ABC gates in international airports, by leveraging multiple frames (see Fig.~\ref{fig:eye}). 
Indeed, we consider the ability to exploit multiple frames is an opportunity to design more accurate and robust MAD systems, enabling for instance the possibility of discarding low-quality frames affected by uneven illumination or non-frontal pose that might harm traditional D-MAD approaches.

From a practical point of view, this study specifically focuses on investigating the effectiveness of providing multiple input frames in current state-of-the-art D-MAD algorithms, including frame-based quality scores and machine learning techniques. By analyzing potential advantages derived from the use of multiple frames over the conventional D-MAD approach, we aim to explore the feasibility and benefits of such an approach in practical scenarios.

Summarizing, our investigation is organized in three sequential steps: 
i) we primarily focus on widely-used score-level fusion strategies across individual frames provided as input to different D-MAD systems~\cite{ferrara2017face, scherhag2020, borghi2021double}; 
then, ii) we analyze the usefulness of face image quality tools~\cite{MagFace,SER-FIQ,CR-FIQA} as an additional input for V-MAD, assuming that image quality metrics can contribute to identifying the most reliable frame for analysis; 
finally, (iii) we exploit machine learning techniques to investigate the potential of artificial intelligence in this task.

By examining these strategies, our goal is to establish a foundation for understanding the potential benefits of leveraging video information in the context of the MAD task, particularly when compared to the classical D-MAD literature approaches.

\section{Related Work}

\subsection{Face Morphing}
Face Morphing is a method of manipulating images whereby one image gradually transforms into another. 
Within the context of electronic Machine-Readable Travel Documents (eMRTDs), this technique enables the creation of facial images that exhibit a double identity. Studies in the literature~\cite{FerraraFM14} indicate that morphed images have the capability to bypass both Commercial-Off-The-Shelf (COTS) FRSs and human controls, rendering face morphing a significant security threat. 
Furthermore, the proliferation of generative Artificial Intelligence techniques, such as Diffusion Models~\cite{Rombach2022CVPR}, Variational Autoencoders~\cite{kingma2019introduction} and Generative Adversarial Networks (GANs)~\cite{goodfellow2020generative}, greatly exacerbates this threat by simplifying the process for potential malicious actors. Additionally, morphed images can be enhanced through either manual or automated retouching procedures~\cite{borghi2021automated, di2024face}, effectively eliminating both detectable and undetectable artifacts. Consequently, there is an urgent need to develop novel MAD systems capable of counterattacking new morphing algorithms and retouching methods. In this scenario, the introduction of the V-MAD paradigm can further improve the accuracy of existing MAD algorithms. 

\subsection{Morphing Attack Detection (MAD)} \label{sec:related_mad}
Since the introduction of the face morphing attack, several approaches have been proposed as potential countermeasures in the literature. 
A recent review of the existing methods is given in~\cite{Scherhag2022} and shows that the research is mainly focused on two different categories of approaches. 
The first one, named Single-image Morphing Attack Detection (S-MAD), relies on a single image, which is analyzed to point out any trace of a possible morphing process. These MAD systems are mainly designed to be exploited during the document enrollment stage, where the ID photo is analyzed for possible inclusion in the eMRTD. 
The second category is Differential Morphing Attack Detection (D-MAD) and includes systems that are supposed to be applied at the face verification stage, such as at airport ABC gates, where the ID photo stored in the document is compared to a trusted live capture acquired at the gate. In this case, two images are available and can be compared for MAD. It is worth noting that only D-MAD approaches can be included in the V-MAD scenario and therefore are introduced and analyzed in the following.

D-MAD methods are based either on traditional computer vision and machine learning techniques or on deep-learning solutions.
One of the current most accurate solutions is proposed in~\cite{scherhag2020}, in which two embeddings extracted through the ArcFace model~\cite{deng2019arcface} are classified by an SVM to decide if the input image is morphed or not. Since the embeddings are obtained using a model trained for the face recognition task, it is possible to infer that the classifier learns to detect the presence of morphing only using information about the subjects' identities.

The idea of exploiting not only information based on identity is proposed in~\cite{borghi2021double}: specifically, the identity features are combined with features related to the presence of visible or not visible morphing traces (artifacts) in the document image. 
In this way, even if two similar subjects are provided as input, the morphing detection ability is preserved.

Other literature methods are not based on learning procedures. In particular, in~\cite{ferrara2017face} a reverse morphing procedure, referred to as demorphing, is used to unveil the genuine identity concealed within the morphed image. 
This approach is based on COTS FRSs; a key challenge arises from the non-linearity inherent in the morphing process, contrary to the linear combination assumed by the authors. Furthermore, the success of the entire process hinges on the accurate estimation of facial landmark positions, with even minor localization errors potentially compromising the efficacy of the entire pipeline.

\begin{figure*}[t!]
    \centering
    \includegraphics[width=0.75\linewidth]{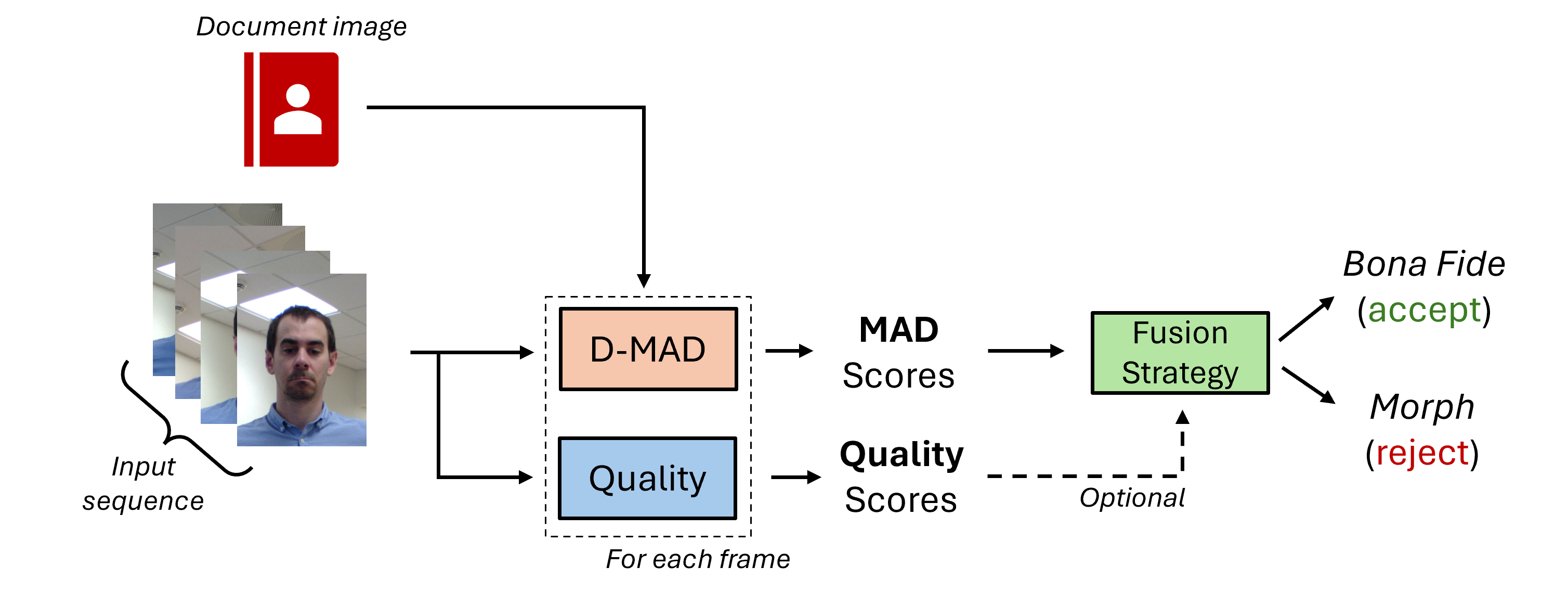}
    \caption{Practical implementation of the V-MAD task. Each frame of the input sequence is analyzed by the same D-MAD model, which receives also the document image as additional input and by a quality tool. Both output MAD and quality scores are then combined through a specific fusion strategy to produce in output the final single score.}
    \label{fig:practical_vmad}
\end{figure*}

\section{Video-based MAD (V-MAD)}
The typical identity verification process at border gates consists of comparing an ID photo $d$ stored into an eMRTD to a sequence of $n$ trusted live-captured frames $\textbf{F} = \left(f_1, f_2, ..., f_n\right)$, acquired for face verification.
Therefore, a theoretical V-MAD system $V(d, \textbf{F})$ should analyze in input the whole sequence $\textbf{F}$ and compare it to the document image $d$ to provide in output a single score to indicate the morphing probability of the document image.   

Being aware that this paper is a seminal work and no V-MAD methods are currently available in the literature, we focus our investigation on adapting D-MAD methods to the V-MAD task, as represented in Figure~\ref{fig:practical_vmad}, to establish a foundation and guidelines for future V-MAD works. 

Therefore, in our V-MAD implementation, we have a D-MAD system $D$ able to compute a morphing score $D(d,f_i)$, representing the probability that the document image $d$ is morphed, based on its comparison with a specific frame $f_i$. This can be repeated for each frame in the sequence of gate images $\textbf{F}$, thus producing a sequence $S(d,\textbf{F})$ of morphing scores: 
\begin{equation}
    S(d,\textbf{F}) = \left(D\left(d, f_i\right), \ \, i = 1, \dots, n\right)
    \label{eq:dmad_scores}
\end{equation}

For the V-MAD task, we consider a MAD system $V$ able to compute a morphing score $V(d,\textbf{F})$ based on the comparison of the document image $d$ with the whole sequence of frames $\textbf{F}$. 
In its simplest form, a V-MAD system can combine, through a function $\phi$, the sequence $S(d,\textbf{F})$ of D-MAD scores (see Eq.~\ref{eq:dmad_scores}) computed on the individual frames $f_i\in \textbf{F}$:
\begin{equation}
    V(d,\textbf{F}) = \phi(S(d,\textbf{F}))
    \label{eq:vmad_from_dmad}
\end{equation}
where $\phi$ is a function $\phi: \mathbb{R}^n \rightarrow \mathbb{R}$, \textit{i.e.,} a function that takes as input a vector of $n$ scores and produces as output a single score for a given sequence.

The function $\phi$, and then the V-MAD task, can be easily generalized to the case where multiple scores are available for each frame $f_i$:
\begin{equation}
    V(d,\textbf{F}) = \phi(S_k(d,\textbf{F}),\ \, k=1, \dots, K) 
\end{equation}
where each $S_k$ is a set of scores computed starting from the document image and the sequence of probe frames $\textbf{F}$. 
Therefore, in this case, the application domain is $\phi: \mathbb{R}^{n \times K} \rightarrow \mathbb{R}$, \textit{i.e.,} $K$ scores available for each of the $n$ frames are condensed in a single output value.

\subsection{D-MAD Score fusion} \label{sec:score_fusion}
Let's focus first on the case where the V-MAD score $V(d,\textbf{F})$ for a given document image $d$ and a sequence of frames $\textbf{F}$ is defined as $S_D(d,\textbf{F})$, applying a function $\phi$ to the D-MAD scores
computed for every single frame $f_i\in \textbf{F}$.

Then, we can define a variety of $\phi$ functions to produce in output a single score as follows:

\begin{itemize}
    
    \item \textbf{Avg}: the average D-MAD score of the sequence $\textbf{F}$
    \begin{equation}
        V\left(d, \textbf{F}\right) = \frac{1}{n} \, \sum_{i = 1}^{n} D\left(d, f_i\right)
    \end{equation}
    \item \textbf{Med}: the median D-MAD score of the sequence $\textbf{F}$
    \begin{equation}   
        V\left(d, \textbf{F}\right) = \underset{i=1,..,n}{\text{med}}D\left(d, f_i\right)
    \end{equation}
    \item \textbf{Vote}: a voting system based on the computed D-MAD scores is defined as follows:
    \begin{equation}   
        V\left(d, \textbf{F}\right) = \frac{1}{n} \sum_{i = 1}^n m\left(D\left(d, f_i\right)\right)
    \end{equation}
    where
    \begin{equation}
        m(D(d, f_i))=
        \begin{cases}
          1 & \text{if} \ D(d,f_i)>thr\\
          0 & \text{otherwise}    \\
        \end{cases} 
    \end{equation}
    \noindent in which $thr$ is a decision threshold.
\end{itemize}

\begin{figure*}[t!]
    \centering
    \includegraphics[width=0.8\linewidth]{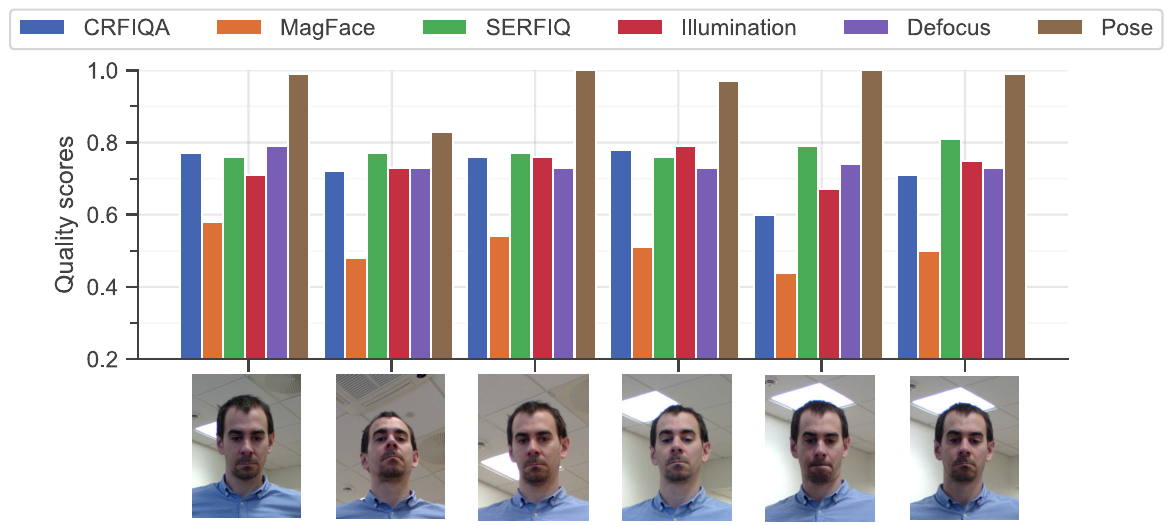}
    \caption{An example of the quality scores computed on a sequence of frames. As reported in Section~\ref{sec:quality}, the first three methods (CRFIQA~\cite{CR-FIQA}, MagFace~\cite{MagFace} and SERFIQ~\cite{SER-FIQ}) are able to compute an overall quality score on the whole image, while the last three are based on specific aspects of the image.}
    \label{fig:quality_on_frames}
\end{figure*}

\subsection{Incorporating Face Image Quality} \label{sec:quality}
A further contribution to the V-MAD task could come from the possibility of exploiting image quality metrics able to assign, for each frame of the probe image sequence $\textbf{F}$, a quality score. 

In this case, the input to the V-MAD model consists of two sets of scores, \textit{i.e.} the D-MAD scores over the single frames ($S_D(d,\textbf{F})$) and the set of quality scores of the probe frames in the sequence ($S_Q(\textbf{F})$): 
\begin{equation}
    V(d,\textbf{F})=\phi(S_D(d,\textbf{F}), S_Q(\textbf{F}))    
\end{equation}
The document image could be considered as well, but since ID document images have to fulfill strict quality requirements \cite{ISO39794}, we expect its quality to be high and will not have a noticeable impact on MAD then we focus on gate images only. 
Two possible $\phi$ functions are considered to combine the single D-MAD scores $S_D=\{D(d,f_i),\ \forall f_i\in \textbf{F} \}$ and the corresponding quality scores $S_Q=\{Q(f_i),\ \forall f_i\in \textbf{F} \}$: 
\begin{itemize}
    
\item \textbf{Weighted average}: the final V-MAD score is computed as the average of the D-MAD scores of each frame, weighted by the corresponding quality score:
    \begin{equation}
        V(d,\textbf{F}) = \sum_{i=1}^n D(d,f_i)\cdot Q(f_i)  
        \label{eq:weighted_avg}
    \end{equation}
    where $Q(f_i)$ is the quality score assigned to the frame $f_i$ by, for instance, a Face Image Quality Assessment Algorithm (FIQAA). 
    \item  \textbf{Best quality}: the V-MAD score is the D-MAD score computed from the frame with the highest quality score:
    \begin{equation}
         V(d,\textbf{F}) = D(d, f_k) \, \text{with} \ k= \underset{i=1,..,n}{\text{arg max}} \, Q(f_i)
         \label{eq:best_quality}
    \end{equation}
\end{itemize}

Even in this case, several algorithms can be used for Face Image Quality Assessment (FIQA); a comprehensive review is available in the recent survey \cite{FaceQualitySurvey}. In general, face image quality can be assessed through a unified score, which takes into account different quality elements and summarizes them into a single value, or by analyzing single quality components related to specific image or face characteristics (e.g., illumination uniformity, blurring, head pose, etc.).
Most of the unified FIQAAs exploit deep learning-based models and have been developed for a wide range of application scenarios where face images are typically acquired in an unconstrained environment and present therefore significant variations. From this category, we consider in particular:
\begin{itemize}
\item \textbf{MagFace}~\cite{MagFace}: a framework proposed for both face recognition and FIQA. It proposes a set of loss functions that learn a universal feature embeddings capable of measuring face quality. Under this new representation, the authors show that the magnitude of the feature embeddings consistently increases for faces more likely to be recognized. MagFace also incorporates an adaptive mechanism to improve within-class feature distributions, ensuring easy samples are pulled closer to class centers while hard samples are pushed away. 
\item \textbf{CR-FIQA}~\cite{CR-FIQA}: a recent method that estimates the face image quality of a sample by learning to predict its relative classifiability, measured according to the allocation of the training sample feature representation in angular space with respect to its class center and the nearest negative class center. To predict the classifiability property of a facial image, the model is trained simultaneously with a face recognition model.
\item \textbf{SER-FIQ}~\cite{SER-FIQ}: the quality assessment score is derived through an unsupervised methodology, relying on the comparative robustness of deeply learned embeddings of the image, rather than on predetermined ground truth acquired from human annotation or facial comparison scores, which may yield imprecise information. This approach analyzes the variability in embeddings generated from random subnetworks of a facial model to estimate the robustness of a sample's representation, and consequently, its quality.
\end{itemize}

In addition to the unified quality scores aimed at an overall evaluation of the face image quality, more specific measures can be used to analyze individual image characteristics. The ISO OFIQ ~\cite{ISO29794-5} standard defines a number of quality components, also providing the guidelines for their computation. We selected here a set of quality components that might have a significant impact on MAD, computed through commercial tools:
\begin{itemize}
    \item \textbf{Illumination uniformity}: it measures the difference in illumination on the left and right sides of the face. It is computed as the intersection of the normalized luminance histograms computed on the left and right parts of the face region, respectively.    
    \item \textbf{Defocus}: it analyzes the level of sharpness. The score is computed as the difference between the face region image and the smoothed version of the same region obtained through a convolution of the image with a mean filter.
    \item \textbf{Pose}: it is focused on the analysis of whether the head pose is frontal. In our analysis, we take into account only the pitch angle since in real operational scenarios yaw and roll angles are mostly well controlled. Only limited variations in pitch might be observed due to the location of the acquisition device at the gate. 
\end{itemize}

An example of these quality scores computed on different frames of a real sequence is reported in Figure~\ref{fig:quality_on_frames}. A visual inspection of the frames suggests that the different quality measures are able to capture to some extent the differences in terms of quality between the images, by assigning lower quality scores when specific issues are visible. 

\begin{figure*}[t!]
    \newcommand{\imagewidth}{0.32}
    \centering
    \begin{subfigure}[b]{\imagewidth\linewidth}
        \centering
        \includegraphics[width=\linewidth]{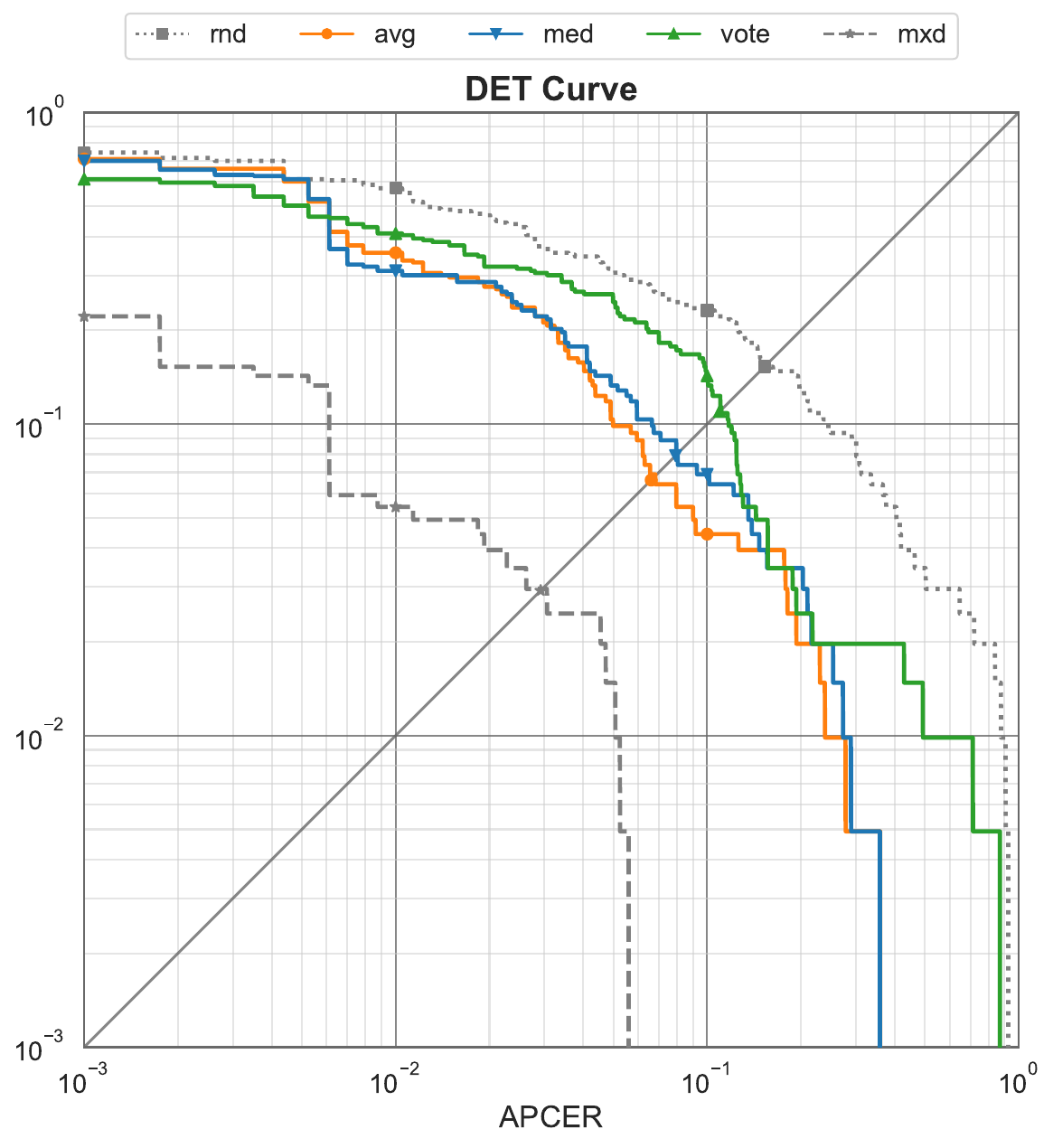}
        \caption{Demorphing~\cite{ferrara2017face}}
        \label{fig:subj-1}
    \end{subfigure}
    \begin{subfigure}[b]{\imagewidth\linewidth}
        \centering
        \includegraphics[width=\linewidth]{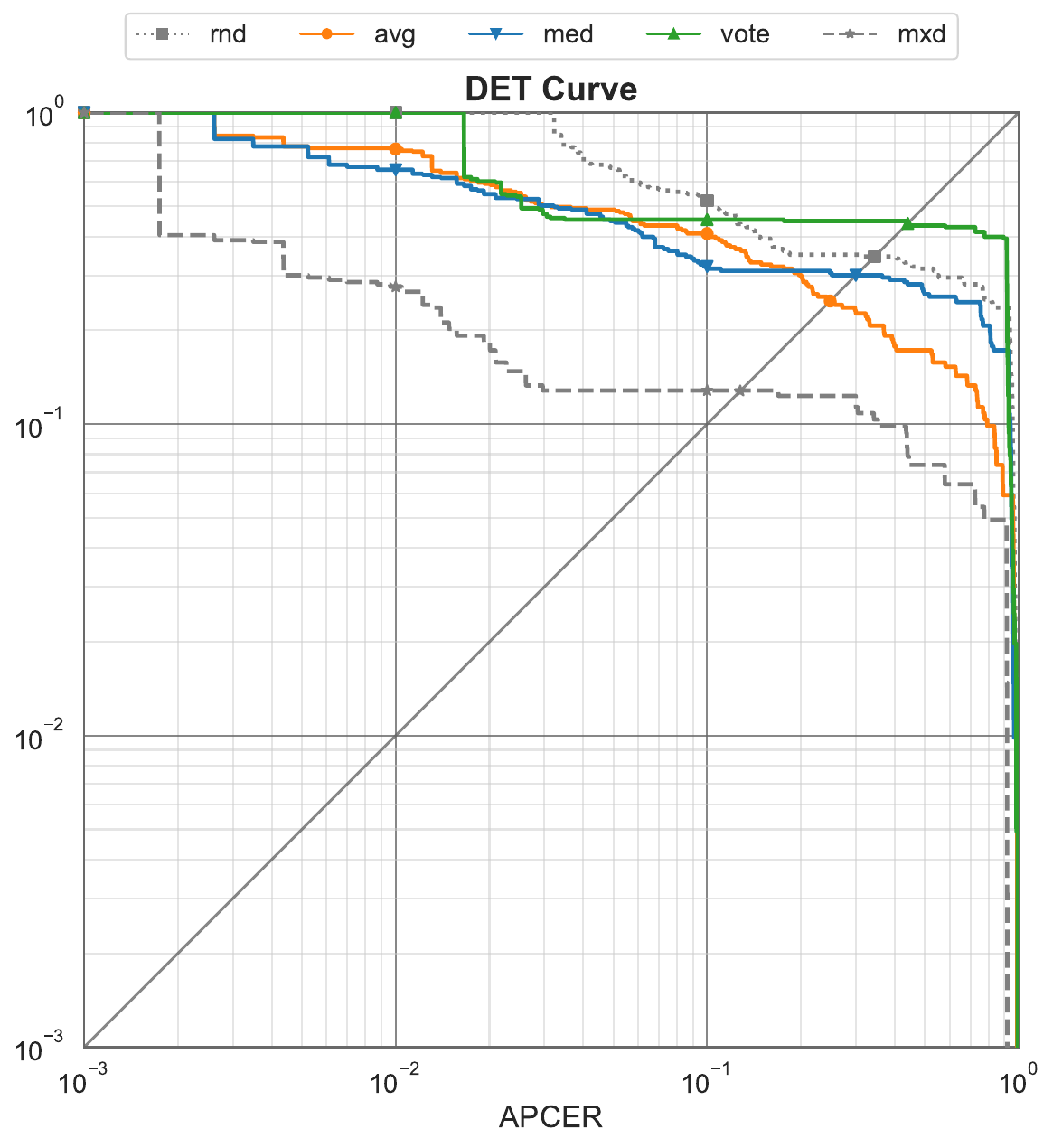}
        \caption{Siamese~\cite{borghi2021double}}
        \label{Siamese}
    \end{subfigure}
    \begin{subfigure}[b]{\imagewidth\linewidth}
        \centering
        \includegraphics[width=\linewidth]{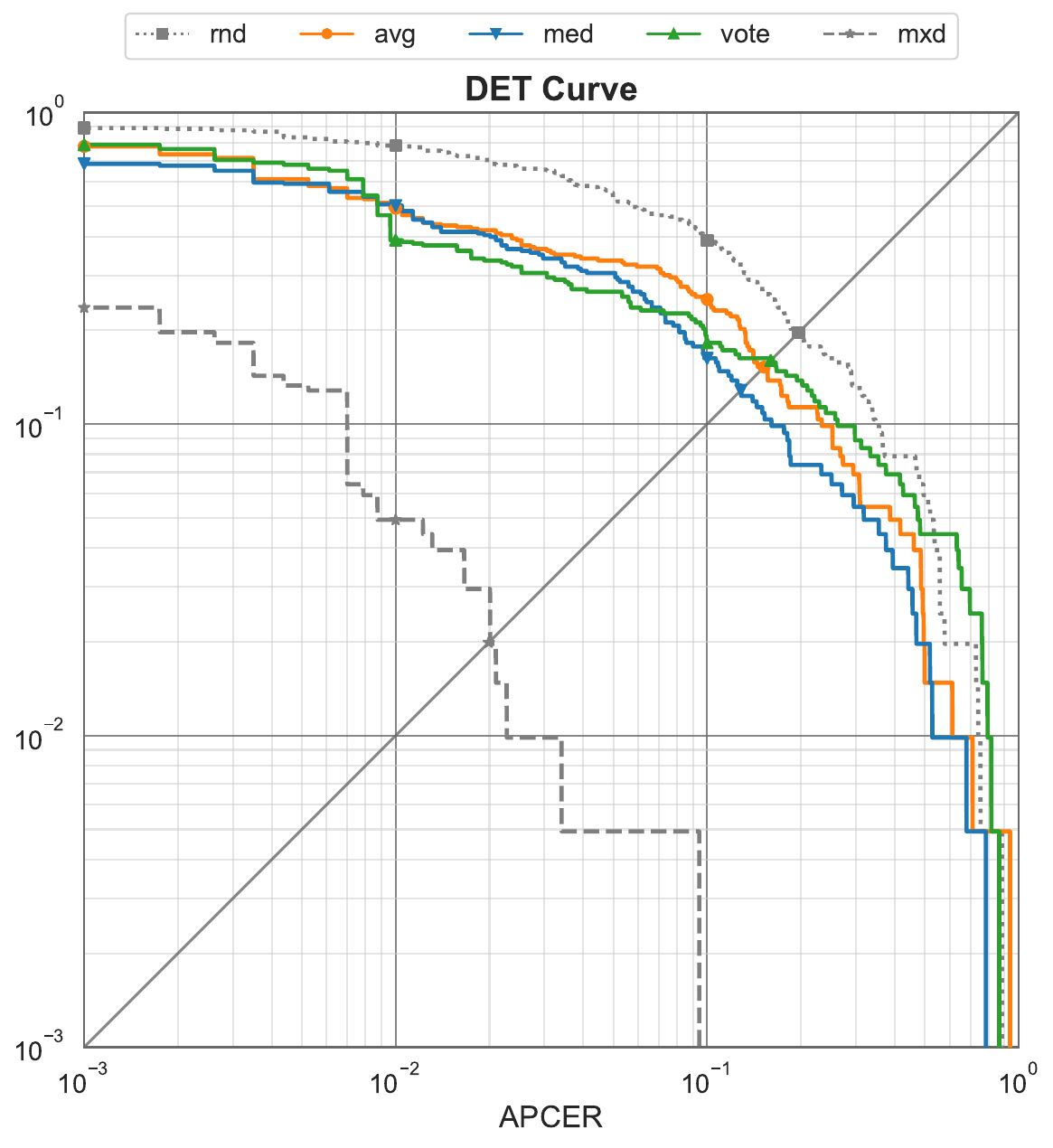}
        \caption{DFR~\cite{scherhag2020}}
        \label{DFR}
    \end{subfigure}
    \caption{V-MAD performance comparison of three D-MAD models using different MAD score fusion strategies (see Sect.~\ref{sec:score_fusion}). The dashed line represents the theoretical upper bound of performance, while the dotted line, based on a random choice in the score set, represents the lower bound. Metrics are expressed as errors, then lower are better.}
    \label{fig:plots_1}
\end{figure*}

\section{Experimental evaluation}
The experimental evaluation is organized in two sequential steps. Firstly, we investigate the impact of the described score fusion strategies (see Sect.~\ref{sec:score_fusion}) applied to scores produced by different D-MAD models. Then, we consider the scores produced by different quality tools (see Sect.~\ref{sec:quality}) in combination with the previous D-MAD scores.

\subsection{Database and evaluation protocol}
Current publicly available datasets commonly used for the MAD task do not well represent the investigated operational scenario, since they do not contain probe sequences.
Therefore, in this paper, a new database has been collected. Images are collected in six different locations, including two airports and four research laboratories, where images were acquired under real border control conditions using authentic ABC gates. A total of 60 different subjects have been involved in the acquisition and some of them have been acquired across multiple locations.

Summarizing, the database contains: 
i) $205$ bona fide document images acquired in a capture setup, which meets the requirements for a document image in a passport application, 
ii) $612$ gate images acquired live with real ABC gates, and 
iii) $1142$ morphed document images created starting from the bona fide document images using $12$ morphing algorithms and various morphing factors.

For the D-MAD task, bona fide document images are compared against gate images of the same subject (for a total of $2187$ bona fide attempts) and morphed document images are compared against gate images of both contributing subjects (for a total of $34698$ morphed attempts). 

For the V-MAD task, bona fide document images are compared against gate sequences of the same subject (for a total of $125$ bona fide attempts) and morphed document images are compared against gate sequences of both contributing subjects (for a total of $1145$ morphed attempts).

\subsection{Metrics}
In the evaluation of the effectiveness of the V-MAD models, we utilize the common error-based metrics tailored to the MAD task as outlined in \cite{raja2020morphing}. 
Specifically, we compute the Bona Fide Presentation Classification Error Rate (BPCER), that quantifies the ratio of authentic images erroneously classified as morphed. 
Conversely, the Attack Presentation Classification Error Rate (APCER) expresses the ratio of morphed images inaccurately identified as genuine.
In the literature, BPCER is often analyzed in conjunction with a predetermined APCER threshold: in our experimental setup, we investigate BPCER$_{10}$ (\textbf{B}$_{10}$), BPCER$_{20}$ (\textbf{B}$_{20}$) and BPCER$_{100}$ (\textbf{B}$_{100}$), denoting the lowest BPCER achievable at APCER values not exceeding $10$\%, $5$\% and $1$\%, respectively. 
It is noteworthy that the latter metric poses a particularly stringent benchmark and conventionally represents the standard operational point for facial verification systems in real-world applications \cite{FrontexGuidelines}.
These values are also visually represented through the Detection Error Trade-off (DET) curve, useful to directly compare different solutions at a glance. 

\begin{table}[h!]
    \centering
    \begin{tabular}{rcccc}
    \toprule
    \textbf{Method}     & \textbf{EER} & \textbf{B$_{10}$} & \textbf{B$_{20}$} & \textbf{B$_{100}$} \\ \midrule
    Siamese~\cite{borghi2021double}     & .392        & .455             & .575              & 1.00              \\ 
    DFR~\cite{scherhag2020}             & .221        & .361             & .486              & .691             \\ 
    Demorphing~\cite{ferrara2017face}   & .150        & .205             & .293              & .501              \\ 
    \bottomrule
    \\
    \end{tabular}
    \caption{Performance of the three different models, on the database used for our evaluation protocol, for the D-MAD task.}
    \label{tab: dmad_original_dataset}
\end{table}

\subsection{Tested D-MAD models}
Following the considerations reported in Section~\ref{sec:score_fusion}, we test different D-MAD methods to produce a MAD score for each frame of a sequence. Specifically, we test three recent D-MAD methods: our implementation of DFR~\cite{scherhag2020} and Siamese~\cite{borghi2021double}, and the official implementation of the method Demorphing~\cite{ferrara2017face}. For each method, further details are reported in Section~\ref{sec:related_mad}.

For the sake of completeness, we report in Table~\ref{tab: dmad_original_dataset} the performance of each model on the evaluation database for the D-MAD task. It is important to note that these results are not directly comparable with the ones obtained in the V-MAD task, but they are useful to understand the performance of the single methods in the simple D-MAD task.
From a general point of view, we observe that the Demorping method exhibits great performance, followed by the DFR (current state-of-the-art method on the BOEP platform), while the effectiveness of the Siamese approach seems to be limited in this scenario.

\begin{figure*}[t!]
    \newcommand{\imagewidth}{0.32}
    \centering
    \begin{subfigure}[b]{\imagewidth\linewidth}
        \centering
        \includegraphics[width=\linewidth]{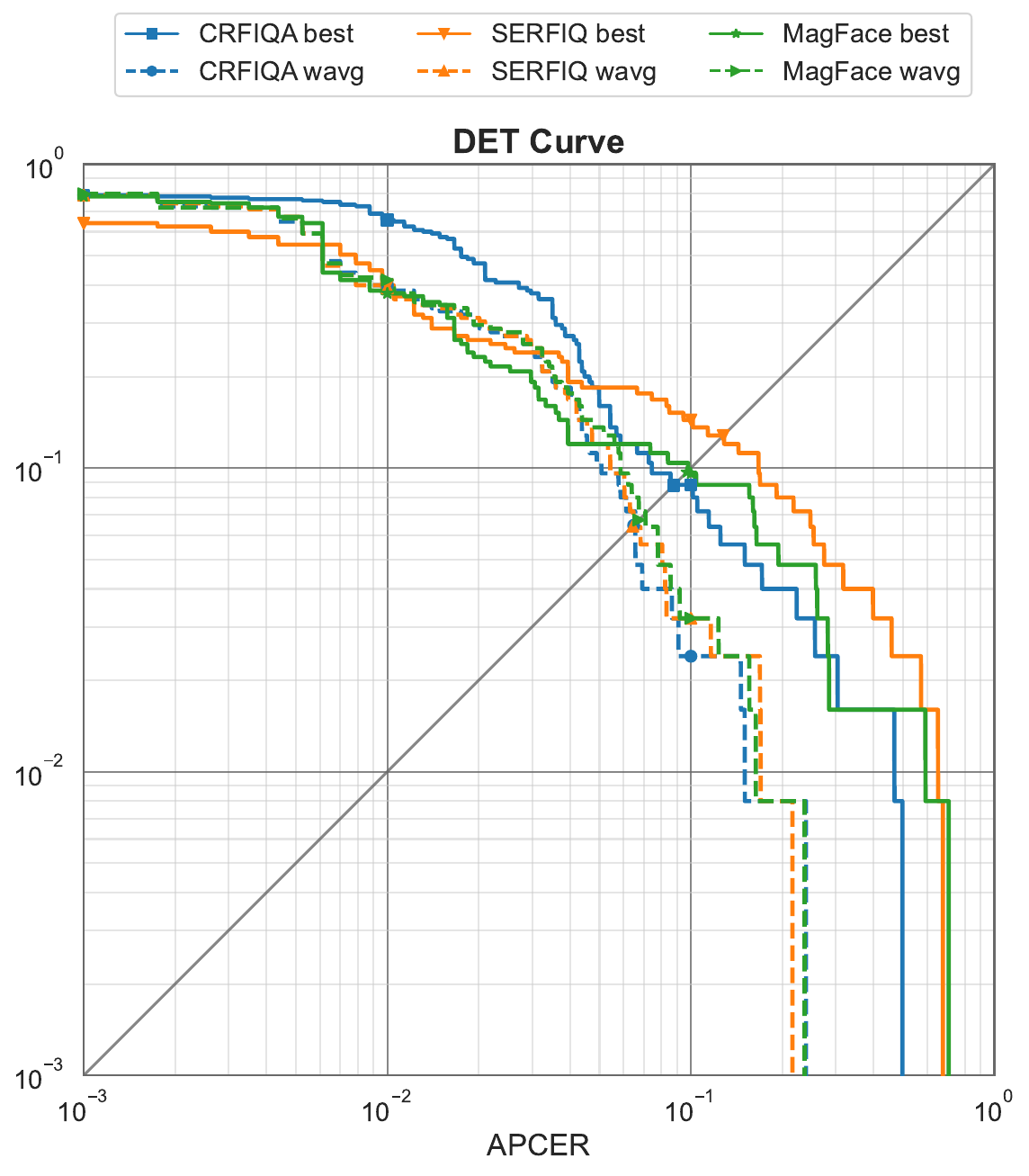}
        \label{fig:subj-1}
    \end{subfigure}
    \begin{subfigure}[b]{\imagewidth\linewidth}
        \centering
        \includegraphics[width=\linewidth]{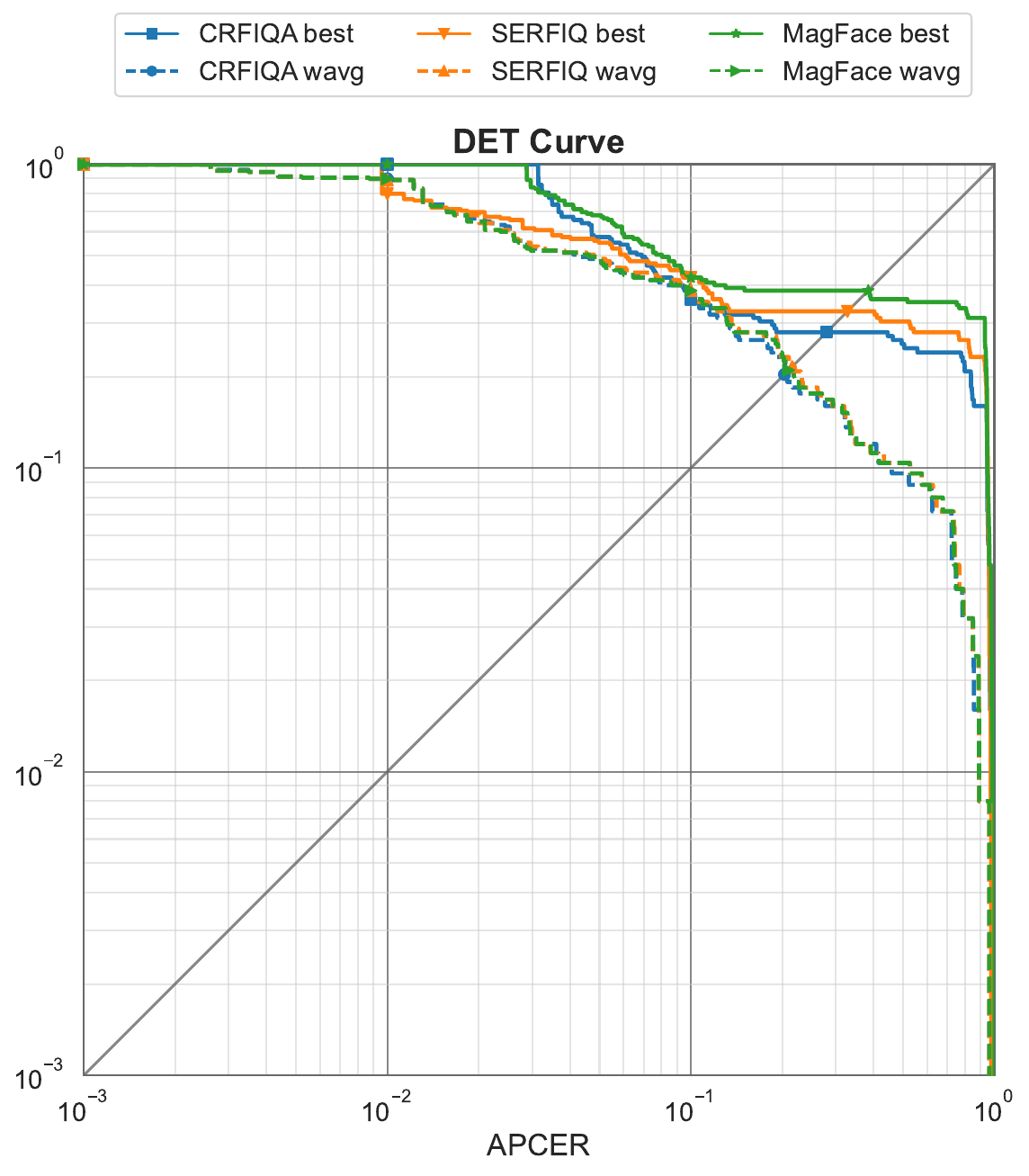}
        \label{Siamese}
    \end{subfigure}
    \begin{subfigure}[b]{\imagewidth\linewidth}
        \centering
        \includegraphics[width=\linewidth]{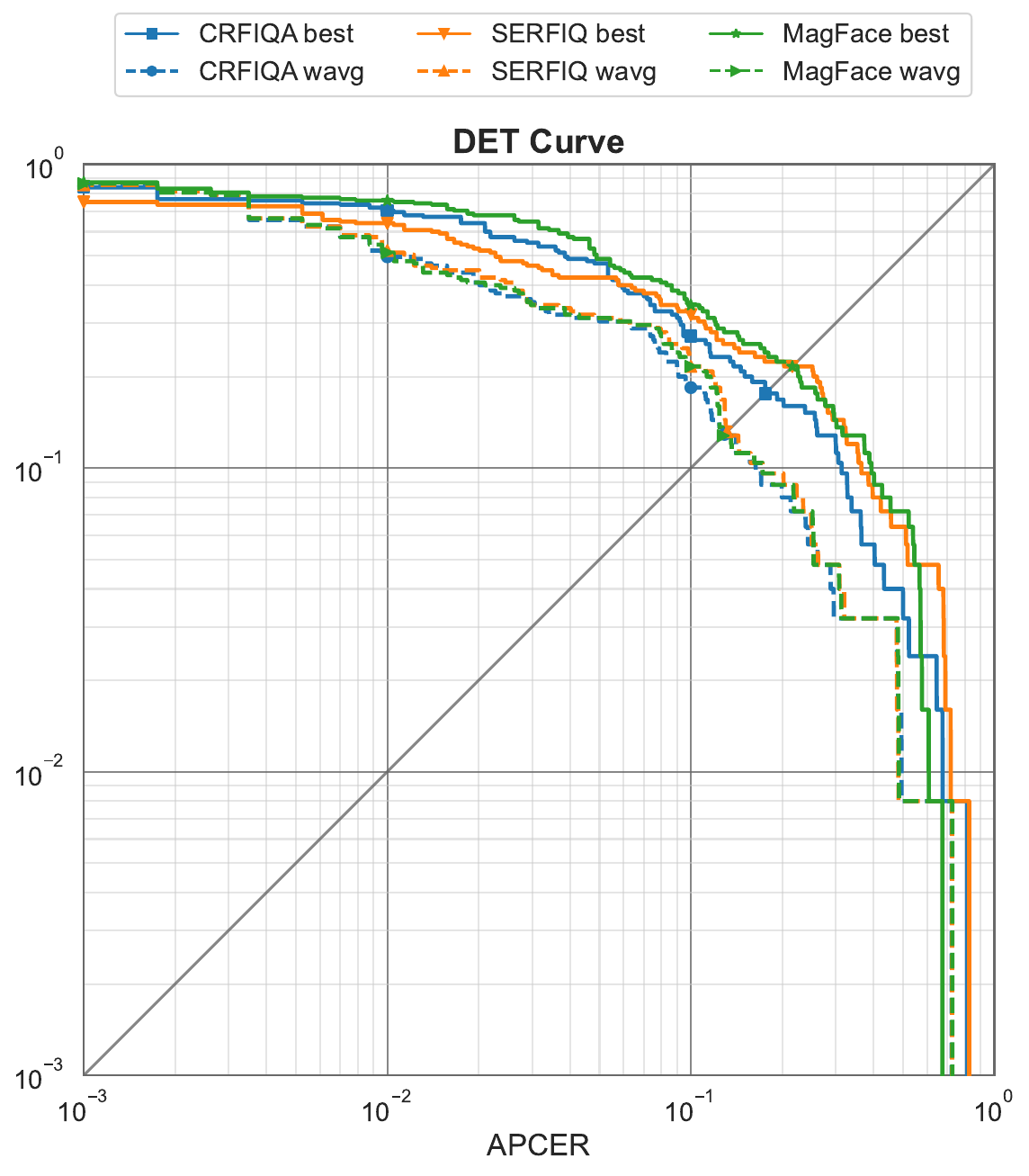}
        \label{DFR}
    \end{subfigure}

        \begin{subfigure}[b]{\imagewidth\linewidth}
        \centering
        \includegraphics[width=\linewidth]{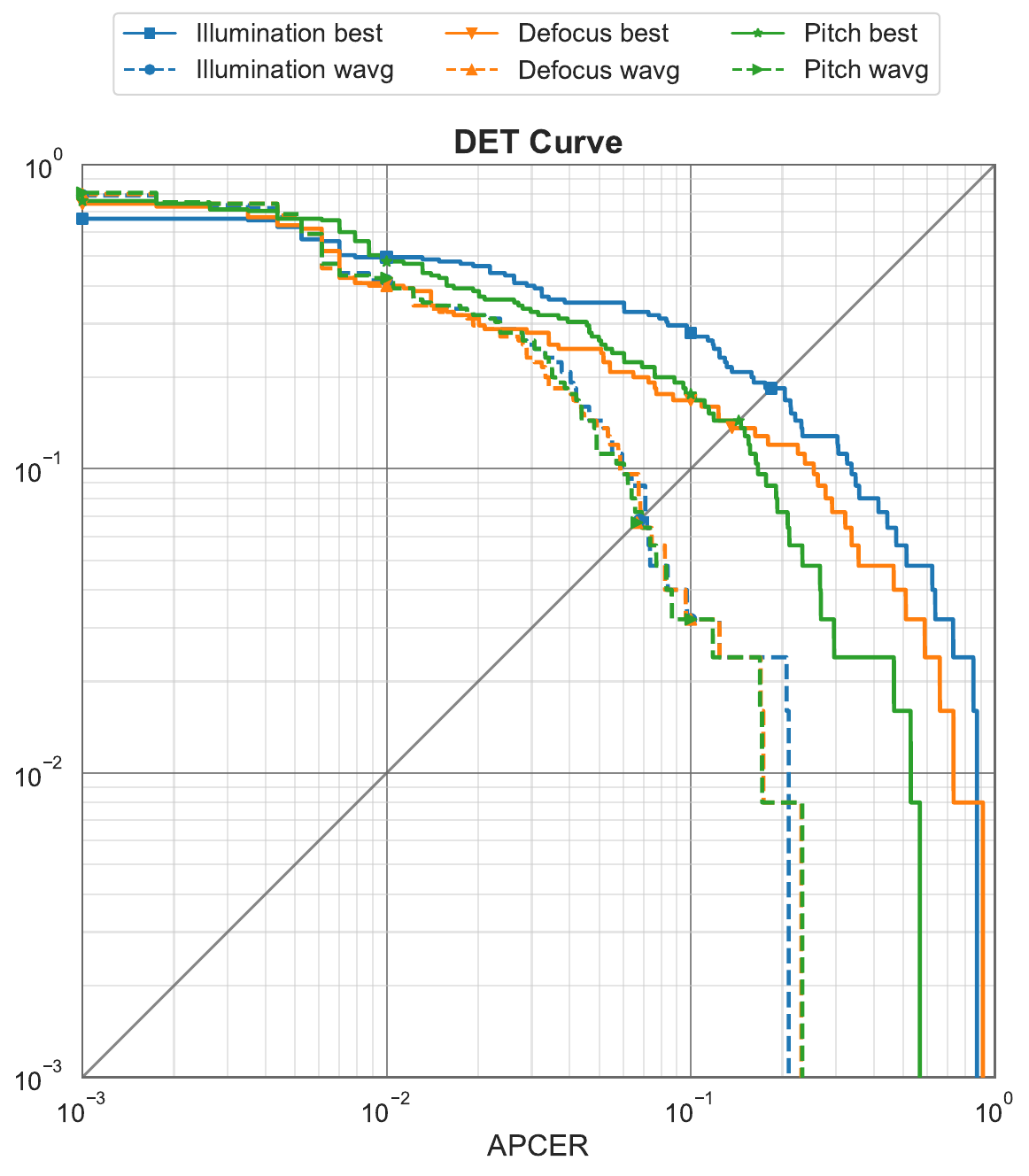}
        \caption{Demorphing~\cite{ferrara2017face}}
        \label{fig:subj-1}
    \end{subfigure}
    \begin{subfigure}[b]{\imagewidth\linewidth}
        \centering
        \includegraphics[width=\linewidth]{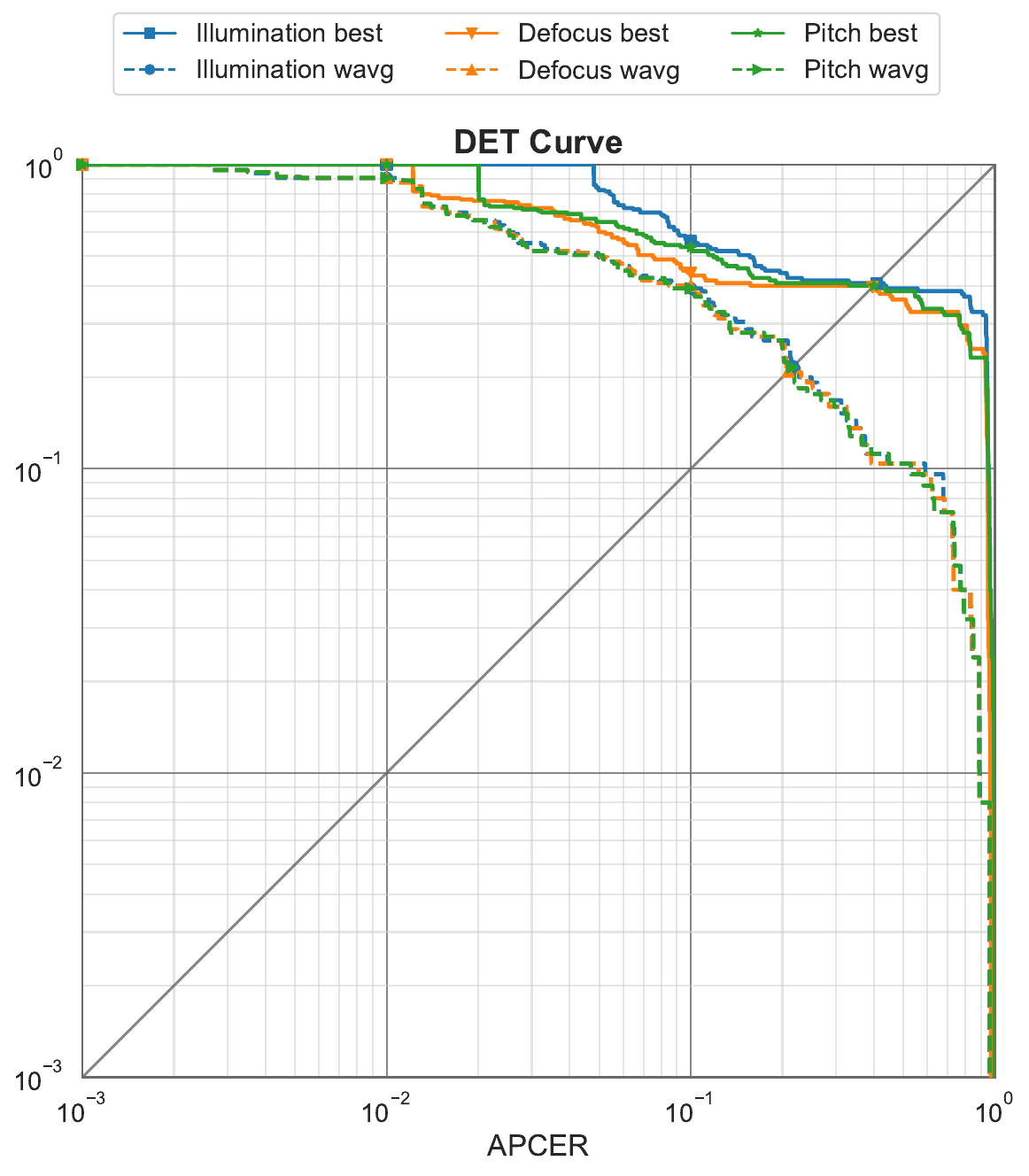}
        \caption{Siamese~\cite{borghi2021double}}
        \label{Siamese}
    \end{subfigure}
    \begin{subfigure}[b]{\imagewidth\linewidth}
        \centering
        \includegraphics[width=\linewidth]{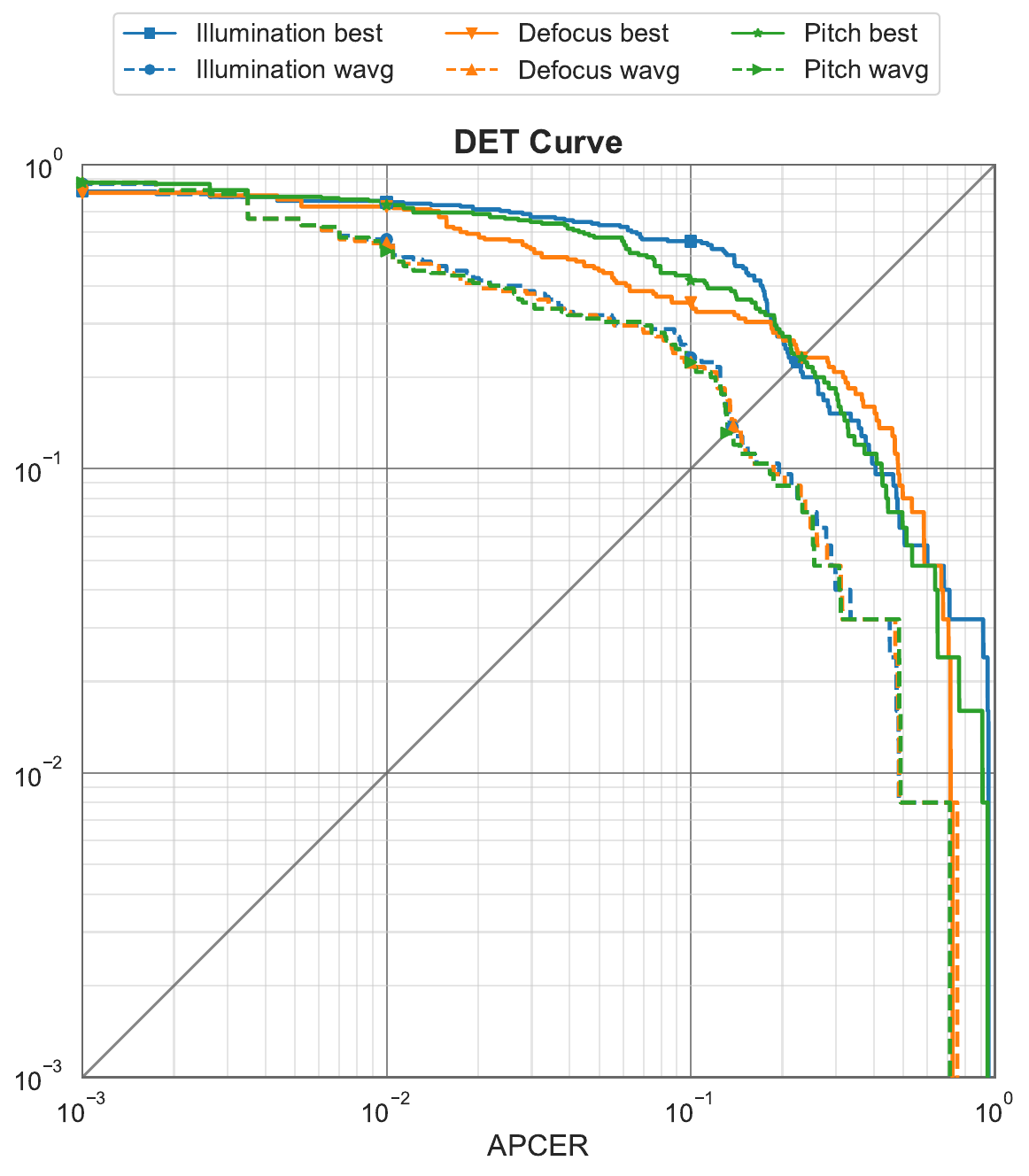}
        \caption{DFR~\cite{scherhag2020}}
        \label{DFR}
    \end{subfigure}
        
    \caption{DET curves for the different V-MAD approaches obtained by exploiting image quality estimated as either unified (first row) or specific (second row) quality measures.}
    \label{fig:plots_2}
\end{figure*}

\subsection{Experimental results}
\subsubsection{Evaluation of D-MAD score fusion strategies} \label{sec:eval_mad}
The results obtained by applying different score fusion strategies to the scores produced by Demorphing \cite{ferrara2017face}, Siamese \cite{borghi2021double}, and DFR \cite{deng2019arcface} are reported in Figure~\ref{fig:plots_1}.
To better understand the range of the performance, we compute two different baselines. The first one, here referred to as ``rnd'' (gray dotted line) is based on the random choice of a single D-MAD score among those available in a given sequence: in this manner, we can compare the performance of each fusion strategy with the corresponding D-MAD approach since we have the same amount of scores for the V-MAD task. 
The second baseline, referred to as ``mxd'' (gray dashed line), simulates an oracle system able to choose for each attempt (either bona fide or morphed) the best possible score. Specifically, we select the minimum or the maximum scores in the given sequence relying on the ground truth annotation: the minimum for the bona fide sequences, and the maximum for the morphed ones. Then, this baseline reveals the theoretical best performance achievable with the scores produced by a specific D-MAD algorithm.

The analysis of the results highlights some important findings.
Firstly, the main observation is that even a V-MAD system consisting of simple score fusion strategies outperforms the tested D-MAD approaches in most cases. 
In other words, in a real scenario, merging the D-MAD scores of multiple frames is better than computing the MAD score on a random frame of the acquired sequence.
In particular, we note that the fusion strategies based on the average or median functions achieve great performance, while the voting system is negatively influenced by the different thresholds to be adopted to compute the votes.
Indeed, the ``avg'' strategy allows to achieve an EER of 0.216, 0.136 and 0.066 for Siamese, DFR and Demorphing, respectively. Even if a direct comparison is not possible due to the different number of attempts, these results give us a clear perception of improvement over the D-MAD results given in Table~\ref{tab: dmad_original_dataset}.

A second important finding is that the ``mxd'' result reveals that theoretically, the V-MAD approach can significantly improve the MAD performance, reaching impressively low EER: 0.048 for Siamese, 0.016 for DFR, and 0.008 for Demorphing.
\subsubsection{Evaluation of the impact of image quality} \label{sec:eval_quali}
Assuming that MAD scores could be estimated more reliably when the quality of the gate image is good~\cite{borghi2021automated}, the second investigation analyzes the impact of combining MAD and quality scores obtained using the approaches described in Section~\ref{sec:quality}. The results of this analysis, reported also in this case through the DET curves, is given in Figure~\ref{fig:plots_2}.

For all the D-MAD methods tested, the results reveal a substantial homogeneity across the different quality models, in particular with respect to the EER.
Moreover, it is possible to note that the weighted average fusion strategy, referred to as ``wavg'' (see Eq. \ref{eq:weighted_avg}) outperforms, even with a limited margin, the best quality strategy (see Eq. \ref{eq:best_quality}). To better appreciate the possible advantages deriving from the use of image quality scores, Table~\ref{tab:comparison_quality_contrib} compares the best results obtained with D-MAD morphing scores only to the best results obtained combining D-MAD and quality scores. For all the D-MAD systems, the introduction of quality has a positive effect since it allows to some extent to reduce the error rates. We only consider here the unified quality scores obtained with MagFace, SER-FIQ and CR-FIQA since they are more effective than single quality components according to the results of Figure~\ref{fig:plots_2}. Among the three FIQAAs, CR-FIQA achieves the best results even if a comparable improvement is observed for the other FIQAAs as well.

\begin{table}[]
    \centering
    \begin{tabular}{crcccc}
    \toprule
    \textbf{Method}      & \textbf{Quality}           & \textbf{EER} & \textbf{B$_{10}$} & \textbf{B$_{20}$} & \textbf{B$_{100}$} \\ \midrule 
    \multirow{4}{*}{Demorphing \cite{ferrara2017face}}  & -                           & .066        & .032             & .120              & .432     \\
                         & {\small CR-FIQA \cite{CR-FIQA}} & .065        & \textbf{.024}             & \textbf{.104}              & .400              \\
    \multicolumn{1}{l}{} & {\small SER-FIQ \cite{SER-FIQ}}                     & \textbf{.064}        & .032             & .120              & \textbf{.392}              \\
    \multicolumn{1}{l}{} & {\small MagFace \cite{MagFace}}                    & .067        & .032             & .136              & .416              \\ \midrule
    \multirow{4}{*}{Siamese \cite{borghi2021double} }     & -                           & .216        & .392             & .504              & .904      \\
                         & {\small CR-FIQA \cite{CR-FIQA}} & \textbf{.203}        & \textbf{.368}             & \textbf{.488}              & .896              \\
    \multicolumn{1}{l}{} & {\small SER-FIQ \cite{SER-FIQ}}                     & .216        & .384             & \textbf{.488}              & \textbf{.888}              \\
    \multicolumn{1}{l}{} & {\small MagFace \cite{MagFace}}                    & .210        & .384             & \textbf{.488}              & .896              \\ \midrule
    \multirow{4}{*}{DFR  \cite{scherhag2020} }         & -                           & .136        & .224             & .312           & .520      \\
                         & {\small CR-FIQA \cite{CR-FIQA}} & \textbf{.127}        & \textbf{.184}             & \textbf{.304}           & \textbf{.496}              \\
    \multicolumn{1}{l}{} & {\small SER-FIQ \cite{SER-FIQ}}                     & .132        & .216             & .312           & .520      \\
    \multicolumn{1}{l}{} & {\small MagFace \cite{MagFace}}                    & .128        & .216             & .312           & .512      \\
    \bottomrule
    \\
    \end{tabular}
    \caption{Comparison of the V-MAD results without (-) and with the contribution of quality scores using the weighted average (wavg) method.}
    \label{tab:comparison_quality_contrib}
\end{table}

\begin{table}[b!]
\centering
\begin{tabular}{cccccc}
\toprule
\textbf{Method}     & \textbf{Input} & \textbf{EER} & \textbf{B$_{10}$} & \textbf{B$_{20}$} & \textbf{B$_{100}$} \\ \toprule
\multirow{4}{*}{Demorphing \cite{ferrara2017face}} & \textit{avg}               & \textit{.065}         & \textit{.016}              & \textit{.127}               & \textit{.381}          \\
           & \textit{wavg}               & \textit{.064}         & \textit{.016}              & \textit{.111}               & \textit{.\textbf{349}}          \\
           \cmidrule{3-6}
                    & $S_D$                 & .063         & .032              & .079               & .524               \\
                    & $S_D$ + $S_Q$               & .\textbf{033}         & .\textbf{000}               & .\textbf{000}                & .460               \\ \midrule
\multirow{4}{*}{Siamese  \cite{borghi2021double} }    & \textit{avg}               & .\textit{209}         & .\textit{317}              & .\textit{460}               & .\textit{968}               \\
                    & \textit{wavg}               & \textit{.206}         & \textit{.302}              & \textit{.460}               & \textit{.968}          \\
                    \cmidrule{3-6}
                    & $S_D$                 & .190         & .222              & .286               & .968               \\
                    & $S_D$ + $S_Q$               & .\textbf{127}         & .\textbf{175}              & .\textbf{254}               & .\textbf{651}               \\ \midrule
\multirow{4}{*}{DFR \cite{scherhag2020} }        & avg               & .\textit{128}         & .\textit{190}              & .\textit{238}               & .\textit{317}               \\
                    & \textit{wavg}               & \textit{.118}         & \textit{.190}              & \textit{.238}               & \textit{.309}          \\
                    \cmidrule{3-6}
                    & $S_D$                  & .111         & .111              & .159               & .206               \\
                    & $S_D$ + $S_Q$               & .\textbf{079}         & .\textbf{079}              & .\textbf{095}               & .\textbf{175}               \\
\bottomrule \\
\end{tabular}
\caption{V-MAD results using a machine learning approach, providing as input the MAD ($S_D$) and quality ($S_Q$) scores.}
 \label{tab:svm_scores}
\end{table}

\subsubsection{Evaluation of the impact of Machine Learning}
A further experiment has been carried out to evaluate if a machine learning approach can be effectively exploited as a fusion strategy. In other words, we investigate if a regressor that receives as input a sequence of MAD and quality scores is able to output a single V-MAD score.

Specifically, we exploit an SVM regressor, with a radial basis function (RBF) kernel, the regularization parameter $C = 1.0$, and the kernel coefficient $\gamma = 10^{-3}$.
In order to have all the sequences, and then the number of the features, of the same length, we empirically set the maximum number of frames to the average sequence length ($50$), padding with null element shorted sequences, and clipping longer ones.
For each frame of the video sequence, we compute the MAD score as reported in Section~\ref{sec:eval_mad}, and quality scores as reported in Section~\ref{sec:eval_quali}. 
Then, both in the training and testing phases, all quality scores are normalized in the range $[0, \, 1 ] $: the scores produced by MagFace are divided by the median value of the entire set of scores of the dataset ($25.77$), while the illumination, defocus and pose scores are divided by $100$ since the original score is in the range $[0, 100]$.
The dataset is split putting the $50$\% of data in training and in the testing sets: therefore, it is important to note that the results of this machine learning approach are not directly comparable with the previous ones reported in Figures~\ref{fig:plots_1} and~\ref{fig:plots_2}, since they are computed on a different portion of the dataset (and then a different number of comparisons).
Different types of classifiers and other combinations have been tested: in the following, we report only the best combinations found based on the SVM model.

Experimental results are reported in Table~\ref{tab:svm_scores}.  
In particular, for each of the D-MAD models investigated, we report three different results. 
In the first line, we report the performance of the best fusion strategy outlined in the first part of our experiments, \textit{i.e.,} the average (avg) and weighted average (wavg) strategies, then without any use of a machine learning approach. 
In the third line, we report the results obtained providing as input to the SVM only the MAD scores, while in the last line, there are the values obtained providing as input the concatenation of the MAD scores, the quality scores produced through the MagFace method and the quality scores related to the illumination uniformity. 
Results suggest that machine learning is a viable approach to merging different scores for the V-MAD task. In particular, SVM overcomes approaches based only on the average and weighted average fusion strategy.

\section{Conclusions}
This study provides significant insights into the effectiveness and potential advantages of the novel Video-based Morphing Attack Detection (V-MAD) task compared to the traditional Differential Image-based (D-MAD) Morphing Attack Detection. 

Firstly, we have demonstrated that incorporating information from multiple frames can lead to substantial improvements in overall performance. Utilizing video sequences enables the development of a MAD system more robust to the inherent variability characterizing face images due to several factors like illumination, pose changes, or motion blur. Even simple score fusion strategies applied to the D-MAD scores computed for the single frames proved to be effective.

Secondly, we proved that face image quality can further contribute to the development of robust V-MAD systems. Unified quality scores as well as single quality components allow to further improve the performance, especially when exploited by means of machine learning models able to combine D-MAD and quality scores into a single effective morphing score.

In conclusion, our study confirms that V-MAD represents a significant evolution from traditional MAD approaches, offering increased effectiveness and robustness in detecting face morphing attacks. Our analysis is a preliminary study aimed at assessing the theoretic feasibility of V-MAD, and the results achieved are still quite far from those of the target oracle system, confirming the need for new and more robust V-MAD systems that, also exploiting the potentiality of deep learning, can effectively work directly on video sequences rather than on single images. The development of new V-MAD approaches will also have to address the issue related to the unavailability of datasets representative of this scenario. Our future research will be dedicated to proposing new contributions to these aspects.

\bibliographystyle{unsrt}  
\bibliography{main.bib}

\end{document}